\documentclass[11pt]{article}

\makeatletter
\renewcommand*{\@fnsymbol}[1]{\ensuremath{\ifcase#1\or *\or 
   \mathsection\or \mathparagraph\or \|\or **\or \dagger\dagger
   \or \ddagger\ddagger \else\@ctrerr\fi}}
\makeatother
\usepackage{acl}
\usepackage{graphicx}
\usepackage{multirow}
\graphicspath{ {./images/} }

\usepackage{times}
\usepackage{latexsym}
\usepackage[T1]{fontenc}
\usepackage[utf8]{inputenc}
\usepackage{arabtex}
\usepackage{utf8}
\setcode{utf8}
\usepackage{microtype}
\usepackage{subfigure}

\title{naab: A ready-to-use plug-and-play corpus for Farsi}

\author{
Sadra Sabouri$^{1 *}$,
Elnaz Rahmati$^{1}$,
Soroush Gooran$^{1}$,
Hossein Sameti$^{1}$ \\
$^1$ SLPL, Department of Computer Engineering, Sharif University of Technology, Tehran, Iran \\
{
sadra@ee.sharif.edu\thanks{\mbox{\ \ } Corresponding author} ,
elnaz.rahmati@sharif.edu,
gooran@ce.sharif.edu,
sameti@sharif.edu
}
}

\begin{document}
\maketitle
\begin{abstract}
The rise of large language models (LLMs) has transformed numerous natural language processing (NLP) tasks, yet their performance in low and mid-resource languages, such as Farsi, still lags behind resource-rich languages like English.
To address this gap, we introduce \textit{naab}, the largest publicly available, cleaned, and ready-to-use Farsi textual corpus.
\textit{naab} consists of 130GB of data, comprising over 250 million paragraphs and 15 billion words.
Named after the Farsi word \RL{ناب} (meaning "pure" or "high-grade"), this corpus is openly accessible via Hugging Face, offering researchers a valuable resource for Farsi NLP tasks.
In addition to \textit{naab}, we provide \textit{naab-raw}, an unprocessed version of the dataset, along with a pre-processing toolkit that allows users to clean their custom corpora.
These resources empower NLP researchers and practitioners, particularly those focusing on low-resource languages, to improve the performance of LLMs in their respective domains and bridge the gap between resource-rich and resource-poor languages.
\end{abstract}

\section{Introduction}
\label{sec:Intro}
Large language models (LLMs) have revolutionized how people interact with technology, representing one of the most significant breakthroughs of the modern era~\cite{srivastava2022beyond,teubner2023welcome}.
While these models have shown remarkable improvements across a wide range of tasks~\cite{chang2024survey} in English, their performance in low and mid-resource languages, such as Farsi, often lags behind~\cite{avetisyan2023large,shen2024understanding}.

Pre-training LLMs to produce Pretrained Language Models (PLMs)~\cite{raffel2020exploring, devlin2018bert} requires vast amounts of data, and large textual corpora are essential for fine-tuning these models for specific languages.
Given this process's time- and resource-intensive nature, having a readily available, large-scale corpus can significantly benefit researchers working to improve NLP in low-resource languages.

The lack of large-scale Farsi text data has made fine-tuning large language models challenging~\cite{habib2021challenges,moniri2024investigating}.
This limitation often restricts the ability to train such models to only a handful of well-funded companies or countries, creating an uneven playing field.
As a result, this lack of accessibility can hinder progress in open science, where collaboration and shared resources are essential for advancing NLP research in these underrepresented languages.

The largest previously available cleaned Farsi textual corpus was a 70GB dataset compiled from eight sources: Common Crawl - fa~\cite{CC-fa}, Miras Text~\cite{Miras}, W2C – Web to Corpus~\cite{majlivs2011w2c}, Persian Wikipedia~\cite{persian-raw-text}, Leipzig Corpora~\cite{biemann2007leipzig}, VOA corpus~\cite{voa-news}, Persian poems corpus~\cite{perisan-poems}, and the Tehran English-Persian parallel corpus (TEP)~\cite{TIEDEMANN12.463}. This corpus has been cleaned and is available for direct download.

Meanwhile, several toolkits have been developed to streamline NLP workflows, including fine-tuning large models.
These tools aim to democratize access to advanced NLP techniques and promote open science.
A notable example is the Python library \verb|transformers|~\cite{wolf2020transformers}, which has become the standard for training and fine-tuning transformer-based models.
Hugging Face has also introduced a range of integrated libraries that enhance accessibility and collaboration in various NLP tasks.

Among these is the \verb|datasets| library~\cite{lhoest-etal-2021-datasets}, which provides open-source corpora that are easily accessible to NLP researchers.
However, none of the existing Farsi corpora are available on \verb|datasets|.
The first contribution of this work is to provide an easily accessible Farsi corpus, available to all through Hugging Face \verb|datasets|.

\begin{table*}[ht]
    \centering
    \begin{tabular}{lcccc}
    \hline
        \textbf{Cleaned Corpus Name} & \textbf{Size (GB)} & \textbf{\# paragraphs} & \textbf{\# words} & \textbf{$\frac{\# words}{\# paragraphs}$}\\ \hline
        Persian NLP & 67 & 13,287,678 & 7,618,898,575 & 573.38 \\
        OSCAR-fa & 36 & 60,099,393 & 4,193,005,807 & 69.76 \\ 
        AGP & 23 & 141,912,688 & 2,776,681,752 & 19.56 \\ 
        LSCP & 2.3 & 15,205,432 & 269,097,323 & 17.69 \\ 
        Telegram & 0.9 & 6,471,586 & 100,253,032 & 15.49 \\
        \hline
        Total & 129.2 & 236,976,777 & 14,957,936,489 & 63.12 \\
    \end{tabular}
    \caption{Statistics of the cleaned corpora included in the \emph{naab} dataset. The table presents the size (in GB), the total number of paragraphs, the total number of words, and the average number per paragraph for each corpus.}
    \label{tab:corpus-stat}
\end{table*}

One of the other primary challenges faced by NLP researchers is effective data pre-processing.
Textual corpora, often derived from web-crawled data, frequently contain undesirable text and personal information.
Traditional methods, such as trimming to remove unwanted patterns~\cite{raffel2020exploring, farahani2021parsbert}, are often computationally expensive and memory-intensive.
In this technical report, we introduce a more efficient alternative: a streaming pipeline for pre-processing texts in Farsi, which addresses these issues in a streamlined and resource-efficient manner.

Our solution to these challenges is embodied in the \textit{naab} project, derived from the Farsi word \RL{ناب}, meaning "pure" or "high-grade"~\cite{abadis}.
The corpus provides 126GB of training data, consisting of more than 224 million sequences and nearly 15 billion words, and 2.3GB of test data, containing nearly 11 million sequences and 300 million words.

The main contributions of this project include the release of the largest cleaned and open-source Farsi corpus, \textit{naab}, hosted on Hugging Face for easy accessibility; and the introduction of an easy-to-use, streaming-based pre-processing approach that enhances efficiency while maintaining high data quality.


\section{Materials and Methods}
This section describes the materials used throughout the project and the methods to prepare the cleaned version of \textit{naab}, now available as an open-source dataset.
The following subsections detail the incorporated base corpora and the pre-processing techniques applied to ensure high-quality data.

\subsection{Base Corpus}
We utilized several corpora to build \emph{naab}, combining various sources of Farsi text to create a comprehensive and diverse dataset.
Table~\ref{tab:corpus-stat} provides statistics such as the size of each corpus, the number of paragraphs, and the word counts.

The largest underlying corpus is the Persian NLP corpus, accounting for 67GB and 7.6 billion words across approximately 13.3 million paragraphs, with an average paragraph length of 573.38 words.
As described in \cite{persian-raw-text}, this corpus aggregates eight separate corpora: Common Crawl (65GB), MirasText (12GB), Web to Corpus (1GB), Persian Wikipedia (787MB), Leipzig Corpora (424MB), VOA corpus (66MB), Persian Poems Corpus (61MB), Tehran English-Persian Parallel Corpus (33MB).
We used a cleaned version of this dataset and further processed it with our proposed preprocessor (see Section \ref{sec:preprocess}).


The two other major base corpora were OSCAR-fa and AGP.
While they contribute significantly to the number of paragraphs (60 million and 141 million, respectively), they exhibit much shorter average paragraph lengths, especially AGP, where the average is only 19.56 words per paragraph.
This reflects a higher volume of shorter, more fragmented text, potentially sourced from informal or social media content.

The AGP corpus, originally a private resource from ASR Gooyesh Pardaz\footnote{\url{http://asr-gooyesh.com/en/}}, has been publicly released through this project.
This corpus contains over 140 million paragraphs, totaling 23GB after cleaning. It is a diverse mix of formal and informal text gathered from websites and social media.

The OSCAR corpus (Open Super-large Crawled Aggregated coRpus)~\cite{abadji2022towards} is a multilingual dataset.
We used the \emph{unshuffled-deduplicated-fa} subset of OSCAR, resulting in 36GB of data after cleaning.

The Telegram corpus, small in size (0.9GB), features shorter paragraphs, averaging 15.49 words per paragraph, further supporting the informal, conversational nature of this dataset. 
Being a popular messaging platform in Iran, Telegram served as a source of informal Farsi text.
We curated a list of channels that span various topics such as news, entertainment, sports, and more.
While smaller in size, this dataset is rich in colloquial Farsi and provides an up-to-date snapshot of daily conversational language.

Lastly, the Large Scale Colloquial Persian Language Understanding (LSCP) dataset~\cite{khojasteh2020lscp}, originally containing approximately 120 million sentences, offers a middle ground between formal and informal content.
We focused on the Farsi portion of the dataset, resulting in 2.3GB of cleaned text comprising 15.2 million paragraphs, with an average paragraph length of 17.69 words.

Overall, the total size of 129.2GB and the diverse word-to-paragraph ratios suggest that \emph{naab} offers a rich blend of formal and informal text types.

\subsection{Pre-process}
\label{sec:preprocess}
We developed a custom pre-processing script that utilizes efficient Linux kernel tools to clean the data with minimal memory overhead.
Unlike traditional methods that load entire datasets into memory for pattern matching — leading to slower processing speeds — our approach pre-processes approximately 1GB of data per minute on an Intel(R) Xeon(R) CPU E5-2699 v3 @ 2.30GHz, offering exceptional speed.

In contrast to memory-intensive approaches that require $O(n)$ memory for large datasets, our method operates in a streaming fashion, reducing the memory requirement to $O(1)$ by processing the data in chunks. This allowed us to pre-process the massive 130GB dataset on a system with only 16GB of RAM.

The script is customizable and available on GitHub\footnote{\url{https://github.com/Sharif-SLPL/t5-fa/tree/main/preprocess}}.
Below, we outline the main steps of the pre-processing pipeline:

\subsubsection{Filtering Non-Farsi Characters}
In the first step, we defined a filter that allows only ``proper'' words to pass the filter through, defined as words containing:
\begin{itemize}
    \item All 32 characters of Farsi (\RL{ا} to \RL{ی})
    \item Arabic characters which are ubiquitous in Farsi (like: \RL{ێ}, \RL{ۇ}, \RL{ۆ}, \RL{ۀ})
    \item Symbolic characters (like '.', '?', '-', ',' and their Farsi version)
    \item Half-space in Farsi ('<200c>')
\end{itemize}

\subsubsection{Unifying Arabic/Farsi Characters}
Many texts use different shapes of Farsi characters interchangeably affecting the text analysis.
To address this issue, less frequent character variants were replaced with their common alternatives.
The substitution rules are listed in Table~\ref{tab:char-subs}

\begin{table}[ht]
    \centering
    \begin{tabular}{|c||c|}
    \hline
       To be substituted & Alternative \\ \hline
        \RL{ێ} | \RL{ي} & \RL{ی} \\ \hline
        \RL{ۀ} | \RL{ة} & \RL{ه} \\ \hline
        \RL{ك} & \RL{ک} \\ \hline
        \RL{إ} & \RL{ا} \\ \hline
        \RL{ڒ} & \RL{ر} \\ \hline
        \RL{ۆ} & \RL{و} \\ \hline
    \end{tabular}
    \caption{Substitution list of characters to their alternative}
    \label{tab:char-subs}
\end{table}

\subsubsection{White Spaces}
After filtering, consequent spaces are normalized to one space (`` '') to ensure consistency. In addition, any empty line resulting from previous cleaning steps is removed.

\subsubsection{Removing Short Lines}
In this final step, lines with fewer than 5 words (controlled by the \verb|MIN_NUMBER_OF_TOKENS| variable) are removed to reduce noise in the dataset.
This ensures that the remaining text is rich in content and suitable for downstream tasks.

\section{Results}
We have released two versions of the \emph{naab} corpus to facilitate various research and development needs.
These versions differ in the level of pre-processing applied, giving users the flexibility to choose based on their requirements.
Both versions are hosted on Hugging Face’s \emph{datasets} library for easy access.

\subsection{naab-raw}
The first version, \textit{naab-raw}, represents the raw, unprocessed compilation of our newly gathered texts combined with existing Farsi corpora.
This version is ideal for users who wish to handle data cleaning and pre-processing themselves, allowing full customization to fit specific tasks and experiments.

We chose Hugging Face's \emph{datasets} library for distributing \textit{naab-raw}, leveraging its ability to efficiently handle large datasets and link to external sources.
Users can access the raw dataset under the repository \verb|SLPL/naab-raw|\footnote{\url{https://huggingface.co/datasets/SLPL/naab-raw}}.
Additionally, we provide a pre-processing script (see Section~\ref{sec:preprocess}) that can be modified to meet different data cleaning and formatting needs.

\subsection{naab}
The second version, \textit{naab}, is a cleaned, ready-to-use dataset that has been pre-processed to remove noise, non-textual elements, and other irrelevant content.
This version is designed for plug-and-play usage, making it especially convenient for users who want to directly get started on model training or fine-tuning without needing to perform additional data preparation.

With \textit{naab}, users can take advantage of Hugging Face's selective download feature, which allows for downloading specific parts of the dataset rather than the entire corpus.
This feature is particularly useful for those working with limited storage or those focusing on a specific subset of the corpus.
The dataset is hosted under \verb|SLPL/naab|\footnote{\url{https://huggingface.co/datasets/SLPL/naab}}, and users can find further information, including detailed usage instructions, in the dataset card.
This version is ideal for researchers and developers seeking a clean, structured corpus for training natural language models, sentiment analysis, or other linguistic tasks in Farsi.

Both versions of the corpus are continuously maintained and updated, ensuring that users always have access to the latest resources.
Additionally, community contributions are encouraged, further enriching the corpus and its potential applications.

\section{Experiments}
\begin{figure*}[ht!]
     \begin{center}
        \subfigure[]{%
            \label{fig:1gram:all}
            \includegraphics[width=0.48\textwidth]{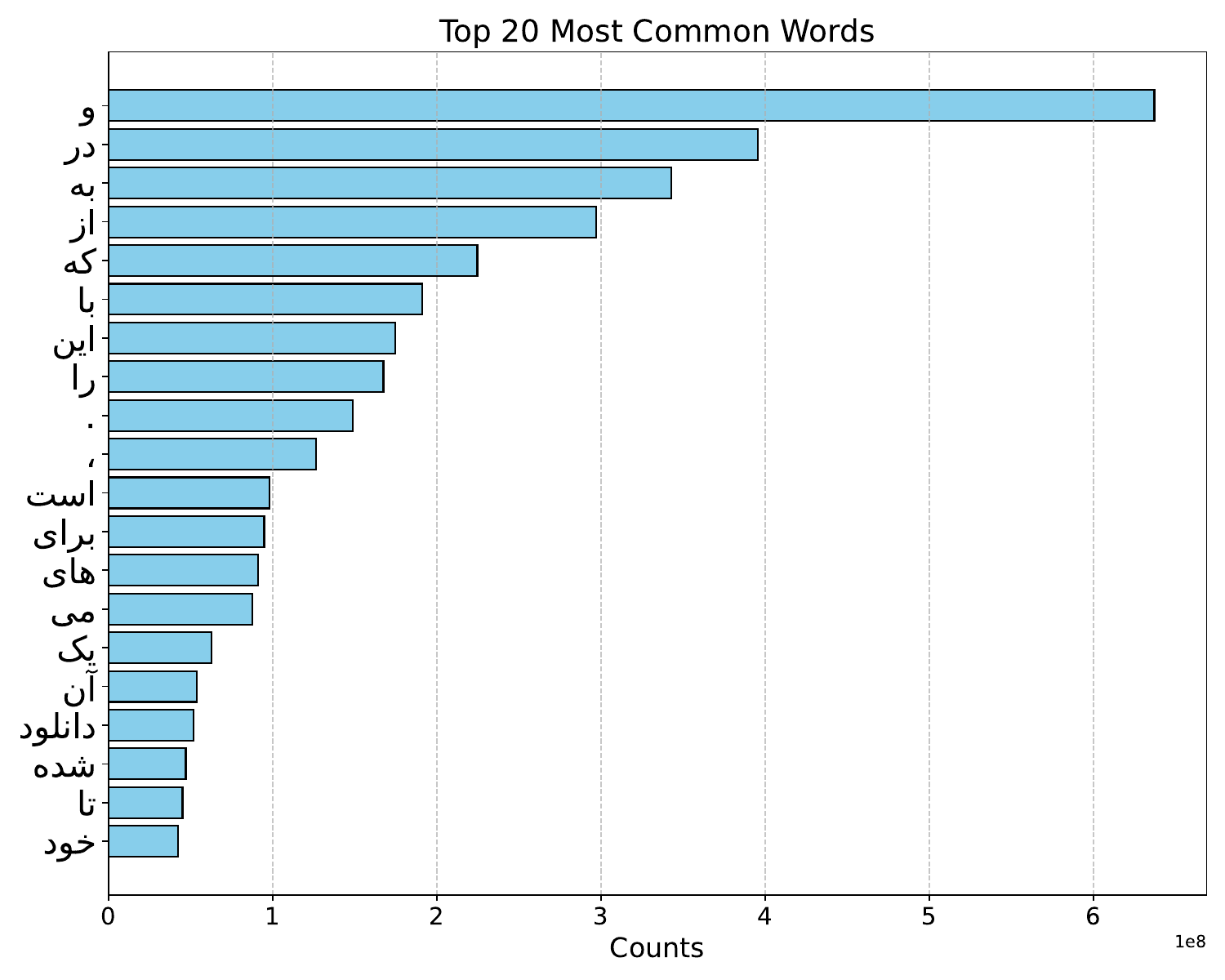}
        }%
        \hfill
        \subfigure[]{%
           \label{fig:1gram:wo_sw}
           \includegraphics[width=0.48\textwidth]{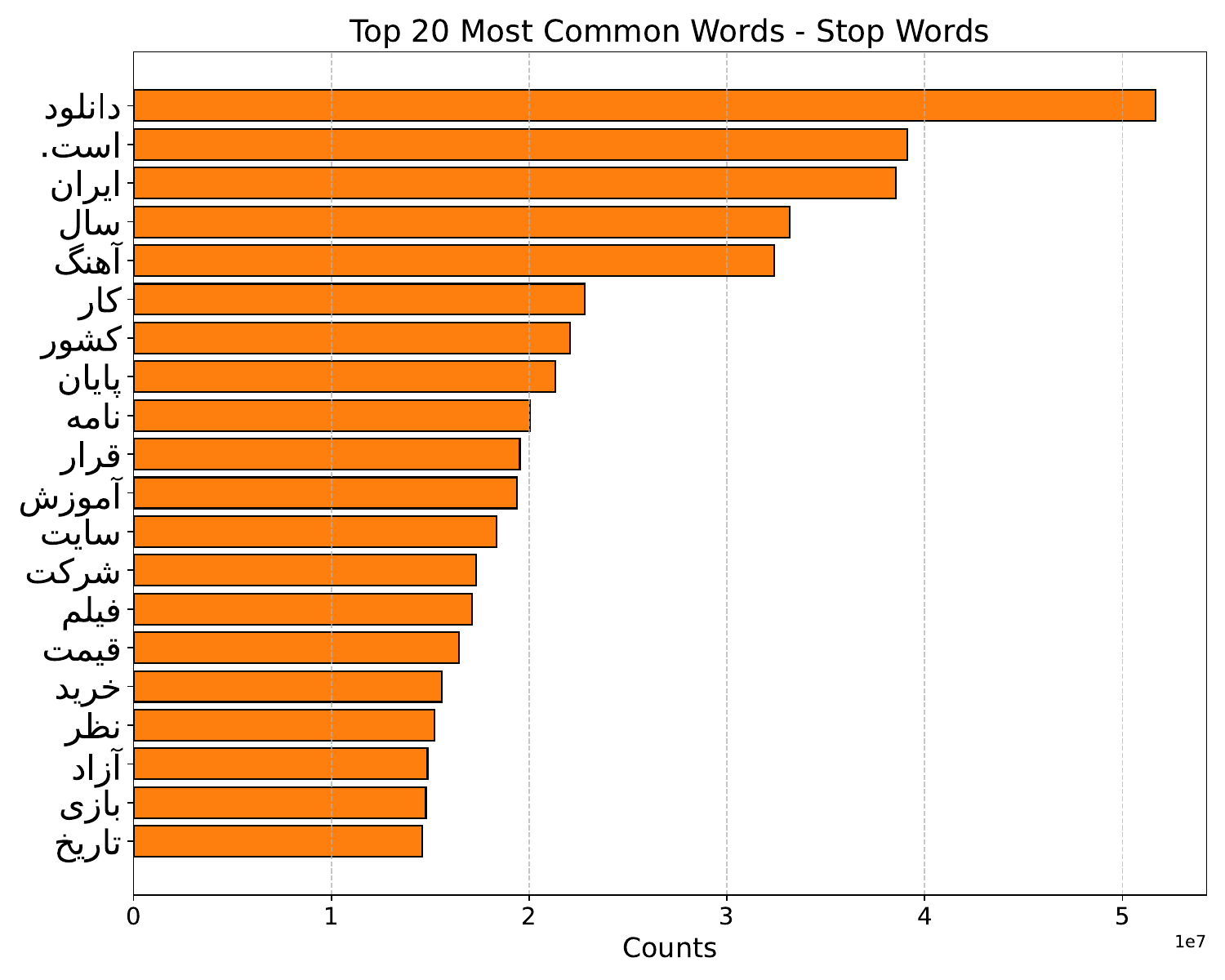}
        }
    \end{center}
    \centering
    \caption{The top 20 most common words in \emph{naab}, along with their corresponding frequencies, are presented in two categories: a) including all words, and b) excluding stopwords. The words are ranked by their counts, with the horizontal axis showing the frequency and the vertical axis listing the words.}
    \label{fig:exp:1grams}
\end{figure*}

We analyzed the frequency distribution of words in the \emph{naab} corpus.
A comprehensive word count was performed, which provides the distribution of the most frequent terms.
This analysis was conducted twice: once including all words in the corpus and once after removing common Farsi stopwords~\cite{faStopWords} to provide a clearer view of the meaningful vocabulary (Figure \ref{fig:exp:1grams}).

The top 20 most frequent words, including stopwords, are visualized in Figure \ref{fig:1gram:all}.
Expectedly, stopwords like ``\RL{و}'' (and) and ``\RL{در}'' (in) dominate the list due to their high frequency in Farsi.
However, after removing stopwords, the most common words shift towards more meaningful terms, as shown in Figure \ref{fig:1gram:wo_sw}.
Once stopwords were removed, frequently occurring terms shifted to content-bearing words.
Notably, words such as ``\RL{ايران}'' (Iran) and ``\RL{دانلود}'' (Download) frequently appeared in the stopword-free list, reflecting the corpus's emphasis on regional topics influenced by internet-sourced content.


\section{Usage \& Future Works}
\label{sec:usage}
Naab corpus serves as an essential resource with broad applications across various domains of Natural Language Processing (NLP) and beyond.
Given its size, diversity, and richness, it offers numerous opportunities for both academic research and practical applications.

One of the primary uses of this corpus is for training large language models (LLMs).
Self-supervised learning approaches, which leverage vast amounts of unlabeled text data, can use this corpus to develop powerful language models.
Traditional models, such as n-gram language models, can use this dataset to capture word sequence patterns and short-term dependencies, providing useful insights into the structure of Farsi.
More advanced transformer-based models, such as BERT~\cite{devlin2018bert}, and BART~\cite{lewis2019bart}, T5~\cite{raffel2020exploring_t5}, GPT~\cite{achiam2023gpt}, Llama~\cite{dubey2024llama}, can also be pre-trained or fine-tuned on this corpus.
With its vast and varied text, these models can significantly improve the understanding and generation of Farsi, making them ideal for a wide range of downstream tasks.

Researchers can use it to develop and improve text classification systems that categorize documents based on topics, sentiment, or other features.
Named entity recognition (NER) models trained on this dataset will be able to detect entities such as names, locations, and dates from Farsi text.
Additionally, part-of-speech (POS) tagging, which assigns grammatical categories to words~\cite{martinez2012part}, can be improved with this data.
Text summarization systems can leverage this corpus to generate summaries of lengthy Farsi documents, making it a useful tool for information extraction and content consumption.

Beyond NLP tasks, this corpus contributes to advancements in speech processing.
Automatic Speech Recognition (ASR) systems, which convert spoken language into text~\cite{hadian2023review,kao2008rapid}, can be trained and fine-tuned using this large collection of Farsi text, improving transcription accuracy.
Text-to-Speech (TTS) models, which synthesize spoken language from text~\cite{kao2008rapid}, will also use the corpus, producing more natural and fluent spoken Farsi.

These technologies are crucial for developing voice-based applications such as virtual assistants and automated customer service in Farsi.

The diversity of the corpus also makes it ideal for various types of linguistic research.
Lexical and semantic studies will benefit from the large-scale dataset, as it provides rich material for exploring vocabulary evolution and usage patterns in Farsi.

Overall, this corpus is not only a critical resource for developing advanced NLP tools in Farsi but also a gateway to exploring the linguistic and cultural richness of the Farsi language.

\section{Conclusions}
\label{sec:Conclusion}
The availability of large-scale textual data is a critical challenge for Farsi language researchers.
In response to this, we present the largest open-source Farsi textual corpus, provided in both a cleaned version, referred to as \emph{naab}, and a raw version, \emph{naab-raw}.
These two datasets are accessible to the research community as open-source resources hosted on Hugging Face’s data hub, making them readily available for a wide range of NLP and linguistic studies.
Furthermore, we introduce a stream-based pre-processing approach, enabling users to efficiently generate their own large-scale text datasets from scratch, accommodating those with specialized processing needs.
This work aims to bridge a significant gap in Farsi language resources, empowering researchers and practitioners to push forward innovations in language technology.

\section{Limitations}
\label{sec:limitations}
\subsection{Stopwords Retained for Contextual Integrity}
In this work, we chose not to remove stopwords to keep the meaningful structure of the text.
Stopwords, while often considered uninformative~\cite{ladani2020stopword}, can play a significant role in understanding the contextual relationships between words in natural language.
However, the inclusion of these high-frequency words may affect certain statistical measures, such as n-gram frequency analysis, and could introduce noise in specific use cases where stopwords are less informative.

\subsection{Duplication Issues}
Due to computational limitations, deduplication was not performed on this corpus.
Given the large size of the text corpus, there is a possibility of duplicated content, as some of the underlying datasets may share the same texts.
This duplication could influence the analysis, and affect the reliability of word embeddings.
Future work could address this by implementing more advanced deduplication strategies to ensure more accurate and unbiased results.
Researchers who wish to use a subset of this dataset can easily perform deduplication, provided they have sufficient computational resources. With manageable dataset sizes, deduplication processes become more feasible.

\subsection{Personal Information}
While we made significant efforts to ensure that the data sources used in the \emph{naab} corpus avoid personal information, the nature of publicly available text data means there is still a potential for personal details to be present.
To minimize this risk, we applied filtering techniques, including removing numerical data that could represent sensitive information like phone numbers, addresses, social security number, and credit cards.
Despite these precautions, some identifiable data may still exist.
Researchers and practitioners using this dataset should be cautious and adhere to ethical guidelines, especially if sensitive information is detected.
It is the responsibility of users to ensure their work complies with all applicable legal, ethical, and institutional standards.
This work and its authors do not take responsibility for any misuse of the dataset, and users are solely responsible for ensuring their usage follows appropriate data privacy protocols.

\subsection{Corrupted Sentences}
Since our pre-processing scripts filtered out specific characters, there is a slight possibility of introducing corrupted sentences in cases where some parts of the input text utilize a different character set or keyboard layout than others.
Although this scenario is highly improbable, it remains a potential issue.
To mitigate this risk, we analyzed part of the texts randomly and found no instances of such corruption.
Nevertheless, we report this as a possible limitation to acknowledge the theoretical risk and ensure transparency.

\section*{Acknowledgement}
\label{sec:Ack}
We would like to thank everyone who has worked hard to promote open science and make resources accessible to all. A special thanks to Mohammadreza Hosseinian, CEO of ASR Gooyesh Pardaz, for kindly allowing us to use their private data and release it as open-source. We thank Sepand Haghighi who provided creative solutions to the challenges we faced and Mehran Ziadloo for his helpful feedback on the pre-print version of this project.

\bibliography{naab}

\end{document}